\documentclass{article}
\usepackage[english]{babel}
\usepackage[utf8x]{inputenc}

\usepackage{apacite}
\usepackage{times}
\usepackage{latexsym}
\usepackage{multirow}
\usepackage{graphicx}
\usepackage{soul}

\title{SEntFiN 1.0: Entity-Aware Sentiment Analysis for Financial News}

\author{Ankur Sinha , 
  Satishwar Kedas ,
  Rishu Kumar \\
  Production and Quantitative Methods,
  IIM Ahmedabad \\
\\
  Pekka Malo \\
  Department of Information and Service Economy \\
  Aalto University, Espoo, FI-00076, Finland\\}

\date{}

\begin{document}
\maketitle
\begin{abstract}
Fine-grained financial sentiment analysis on news headlines is a challenging task requiring human-annotated datasets to achieve high performance. Limited studies have tried to address the sentiment extraction task in a setting where multiple entities are present in a news headline. In an effort to further research in this area, we make publicly available SEntFiN 1.0, a human-annotated dataset of 10,753 news headlines with entity-sentiment annotations, of which 2,847 headlines contain multiple entities, often with conflicting sentiments. We augment our dataset with a database of over 1,000 financial entities and their various representations in news media amounting to over 5,000 phrases. We propose a framework that enables the extraction of entity-relevant sentiments using a feature-based approach rather than an expression-based approach. For sentiment extraction, we utilize 12 different learning schemes utilizing lexicon-based and pre-trained sentence representations and five classification approaches. Our experiments indicate that lexicon-based n-gram ensembles are above par with pre-trained word embedding schemes such as GloVe. Overall, RoBERTa and finBERT (domain-specific BERT) achieve the highest average accuracy of 94.29\% and F1-score of 93.27\%. Further, using over 210,000 entity-sentiment predictions, we validate the economic effect of sentiments on aggregate market movements over a long duration.
\end{abstract}

% \begin{document}
% \maketitle

\section{Introduction}
Financial sentiment analysis (FSA) addresses the problem of \textit{automatically} extracting sentiments from financial text such as documents, stock message boards, news headlines etc. \cite{kearney2014textual} with minimal human intervention. Using FSA systems, market participants seek to gain edge over competitors towards profitable investment or trade decisions. Of significant interest is the sentiment analysis of financial news flow which is a timely and reliable source of information on market events. The news flow presents novel information to the financial market participants\textit{, for ex.} enabling the traders to conduct company or industry directed trades \cite{von2015first} and investors to actively manage portfolios \cite{seo2004financial}. Over the last decade, FSA has also attracted significant research attention driven by the abundance of data, the utility of sentiment information, and the nuances of sentiment bearing expressions \cite{das2001yahoo, malo2014good, van2015good}. In our study, we focus on fine-grained FSA, in which the task is to extract sentiments for the identified entities in the financial text \cite{schouten2015survey}. Our sentiment classification framework is directed towards the extraction of sentiment towards the identified entities in the news text.

%\st{Financial news flow is a key source of financial and economic information for investors, analysts, and traders that can lead them to action.  \st{among the key uses}. By evaluating the sentiments expressed in news headlines, market participants aim to gain an edge over others in making profitable investment decisions \st{with respect to financial instruments such as stocks, bonds, and derivatives}. In a highly competitive industry, even the slightest amount of performance gains can lead to large profits.  Fine-grained sentiment analysis which aims to achieve higher performance, focuses on the task of identifying sentiments for entities or aspects present in the financial text and utilizes specialized finance-focused lexicon and datasets \cite{schouten2015survey}.}

%  and typically the entity-recognition task uses a Named Entity Recognition (NER) system

Fine-grained sentiment analysis or aspect-based sentiment analysis (ABSA) entails two different tasks - the recognition of entities (also aspects or topics) and the extraction of sentiments corresponding to the identified entities. Researchers in fine-grained FSA often focus exclusively on the sentiment extraction task, often using Named Entity Recognition (NER) systems for entity-recognition or utilizing entity-annotated datasets. In the context of financial news flow, entities are often the companies or organizations towards which the news text expresses sentiment. Research in financial sentiment extraction task has led to three streams of literature - a) sentiment lexicon \cite{wilson2005recognizing, loughran2011liability, malo2013learning}, b) sentiment-annotated datasets such as the JRC Corpus \cite{balahur2013sentiment},  Financial Phrase Bank \cite{malo2014good},  SemEval 2017 Task 5 SubTask 2 Headlines Dataset \cite{cortis2017semeval} and Financial Opinion Mining and Question Answering (FiQA) 2018 Task 1 Sentiment Scoring Dataset \cite{maia201818}, and c) annotation schemes such as the polarity expressions \cite{van2015fine, van2015good} and MPQA Opinion Corpus \cite{deng2015mpqa}. With the advent of language models based on deep learning architectures the focus has shifted from dictionary-based approaches to dataset-driven approaches. 

%In our study, we focus on entity-aware sentiment classification, in which the sentiment classification task is directed towards the entity identified in the news text.

%\st{Research in fine-grained sentiment analysis has led to the development of resources such as finance-focused dictionaries \cite{wilson2005recognizing, loughran2011liability, malo2013learning} and datasets such as the JRC Corpus \cite{balahur2013sentiment},  Financial PhraseBank \cite{malo2014good}, MPQA Opinion Corpus \cite{deng2015mpqa}, SemEval 2017 Task 5 SubTask 2 Headlines Dataset \cite{cortis2017semeval} and FiQA 2018 Task 1 Sentiment Scoring Dataset \cite{maia201818}.} 

Most of the financial news datasets, unfortunately, are only effective in cases where there is a single entity in the news headlines. FSA systems trained on these datasets fail to achieve high performance in the presence of multiple entities especially with conflicting sentiments in a news headline. Considering the limitations of the existing datasets, through our work, we make publicly available SEntFiN 1.0, a human-annotated financial news dataset containing 10,753 news headlines annotated for the entities present and their corresponding classes (positive, negative, and neutral, hereafter referred to as sentiments) with an average sentence length of 9.91 words. The dataset contains 2,847 headlines with at least two entities and 1,233 headlines containing conflicting sentiments. To our best knowledge, SEntFiN 1.0 is the largest publicly available financial news dataset annotated for multiple entities and their corresponding sentiments.

\begin{figure*}[h]
\centering
	\includegraphics[width=\textwidth,keepaspectratio]{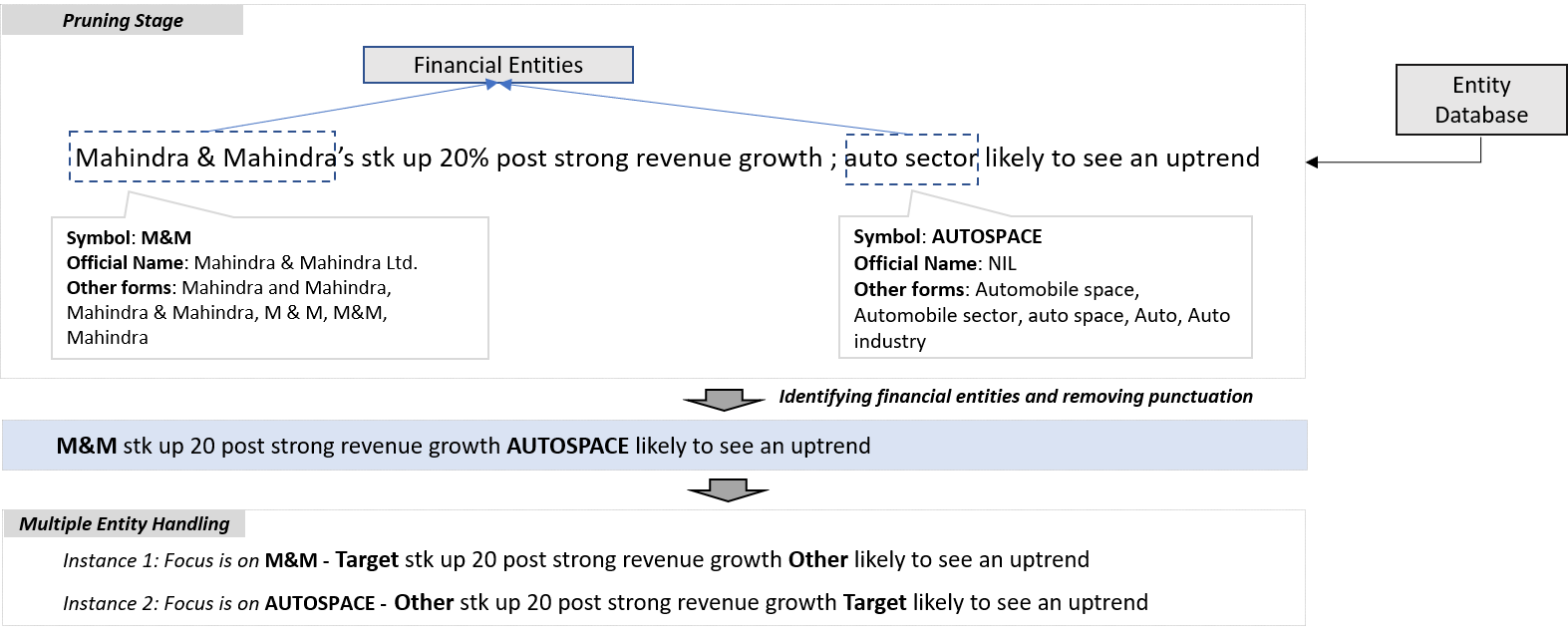}
	\caption{The news headline is first processed to recognize the relevant entities using the entity database. The entity recognition step is performed prior to the pre-processing step. Pre-processing step includes the removal of punctuation and special characters. Handling of multiple entities using two features - Target and Other, with the entity in focus replaced with the Target feature and the entities not in focus replaced with the Other feature. Each news headline can lead to multiple instances of text each with a different Target-Other feature annotation.}
	\label{fig:me_handle}
\end{figure*}

Sentiment analysis based on news headlines is a difficult task. News headlines contain high signal content and have a low risk of spurious information; however, they often have short text span. If multiple entities are mentioned in the news headlines, the syntactic structure capturing the interactions between the entities and their sentiment bearing expressions is required for effective sentiment extraction. The machine-learning based NER systems currently available for public use such as the Stanford CoreNLP NER \cite{manning2014stanford} are trained on multiple corpora - the CoNLL 2003 \cite{sang2003introduction}, MUC-6 \cite{grishman1996design}, MUC-7 \cite{chinchor1997muc}. As such, they suffer from two primary limitations- first, the models are probabilistic in nature and therefore the errors in entity recognition can lead to wrong economic decisions, and second, the training data suffers from temporal and geographical limitations and therefore may not support extensive economic analysis. The annotation schemes such as polarity expressions are limited in their scalability and the NER system (Stanford CoreNLP NER) indicated multiple errors in recognizing entities and entity spans as discussed in Section~\ref{sec:entitydatabase}. Therefore, to conduct our experiments on identifying economic value of news-based sentiments, we developed an entity database covering 1,009 financial entities (companies,sectors) and their various forms of representations in news media amounting to a total of 5,070 phrases. Using this entity database, we were able to recognize the entities mentioned in the news headline without errors. We address the task of recognizing relevant entities using a method of annotating entities with two features - \textbf{Target} and \textbf{Other} resulting in multiple instances of the same sentence with different feature annotations. In each instance, the entity for which the sentiment needs to be extracted is replaced with \textbf{Target} and the other entities are replaced with \textbf{Other} as described in Figure~\ref{fig:me_handle}. With these features, we utilize the sentence as it is, and avoid the use of dependency parsing, part-of-speech tagging, polarity expression, and other methods used in prior studies \cite{ding2014using, malo2014good, van2015good}. Our entity database adds value by providing a list of financial entities and their entity spans as officially recognised, and indicates the requirement of developing specialized NER systems for use in finance.

In the SemEval 2017 SubTask 4 on Sentiment Analysis in Twitter, systems that utilized Deep Learning based architectures such as Recurrent Neural Networks (RNN) \cite{rumelhart1986learning} and Long-Short Term Memory (LSTM) \cite{hochreiter1997long, graves2005framewise} with pre-trained word-embeddings such as GloVe \cite{pennington2014glove} were placed among the top performing systems \cite{moore2017lancaster, ghosal2017iitp, cabanski2017hhu, mansar2017fortia}. Further, recent research utilizing pre-trained context-based representations of text such as the Bidirectional Encoder Representations from Transformers (BERT) \cite{devlin2018bert} fine-tuned to the financial domain using Financial Phrasebank \cite{malo2014good} outperform previously best performing models in sentiment analysis \cite{araci2019finbert, hiew2019bert}. In this work, we utilize three models based on BERT - finBERT \cite{araci2019finbert}, DistilBERT \cite{sanh2019distilbert}, and RoBERTa \cite{liu2019roberta}. Overall, we evaluate 12 different learning schemes and the results are reported in Section~\ref{sec:learning-schemes}. As baselines, we utilize the Linearized Phrase Structures (LPS) representations \cite{malo2014good} with Support Vector Machines (SVM) and Gradient Boosting Machines (GBM). Our experiments indicate that RoBERTa achieves highest average accuracy of 94.29\% and finBERT, a domain-specific BERT model achieves the highest average F1-score of 93.27\%. Lexicon-based N-gram ensemble representations with Multi-Layer Perceptrons (MLP) can achieve above par performance of GloVe-based representations with deep neural architectures. To further the claim that the entity agnostic sentiment analysis is not sufficient for sentiment extraction, we provide a comparison of the proposed entity-based sentiment extraction method against the non-entity-based sentiment extraction approaches, like, NLTK Vader and HuggingFace Sentiment system with Twitter-roBERTa-base model in Section \ref{sec:results}.

We utilize the best-performing sentiment extraction model identified through our experiments to extract the sentiments from over 470,000 news headlines from ``The Economics Times'' on a daily basis from the years 2002 to 2017\footnote{Available upon request from the corresponding author}. We construct an aggregate market sentiment index in correspondence to the broad-based market equity index NSE 500, and study the predictive value contained in the sentiments. In the duration 2012 to 2017, our experiments indicate that the sentiments of the news flow generated in the after-market hours, have an impact on the after-market log price returns and percentage price returns of the NSE 500 index. We observe that the relationship holds significant across multiple years. Our study highlights the sentiments as potential underlying mechanisms through which market participants exchange and act upon information related to market events. We conclude our study with a discussion on the potential future research directions. The study also holds implications for commercial news-based sentiment analysis solution providers such as Bloomberg, Thomson Reuters, and RavenPack whose sentiment analysis systems are designed for high speed, high volume, large variety, and best accuracy. At high speeds, due to the large volumes of news flow on a large basket of financial entities, it is beneficial to operate with short news text for the reasons of scalability and efficiency, and our proposed and validated framework serves as a robust approach for extracting entity-specific sentiments from news. 

To summarize, our contributions are three-fold:
\begin{itemize}
\item We build and release the SEntFiN 1.0, the largest human-annotated financial news dataset, annotated for multiple entities in the news headlines and their corresponding sentiments, containing 10,753 news headlines with 14,404 entity-sentiment annotations. We also provide an entity database of over 1,000 financial entities and their recognized entity spans in news media aiding the entity recognition task.

\item We conduct multiple experiments comparing 12 learning schemes based on two different forms of sentence representations - lexicon-based vs pre-trained. Our findings indicate that RoBERTa outperforms all other learning schemes by a wide margin and that certain lexicon-based approaches achieve a comparative performance to deep-learning approaches.

\item Our experiment on sentiments accumulated during the after-market hours and their effects on opening market prices, confirms the hypothesis that news sentiments hold economic value, and the relation holds over the long-term. We also provide access to over 470,000 news headlines with timestamps, entities, and sentiments pertaining to the Indian Economy.
%We propose an experiment to evaluate the economic value of sentiments based on the knowledge that market price movements occur only during the market open hours and news accumulation occurs throughout the day. 

\end{itemize}

The rest of the paper is structured as follows. In Section~\ref{sec:related-work}, we discuss the related work on the annotated financial news datasets, the financial sentiment lexicon, and the applications of deep learning approaches and lexicon-based approaches for sentiment extraction. In Section~\ref{sec:dataset}, we introduce SEntFiN 1.0, the sentiment-entity annotated financial news dataset that we release through this paper, and use for our experiments and discuss the annotation process, inter-annotator agreement, and dataset statistics. Followed by the discussion on the learning schemes, experiments, and results of our modelling approaches in Section~\ref{sec:learning-schemes}. In Section~\ref{sec:informationContent}, we discuss the construction of our experiment and the evaluation of hypothesis on the economic value of sentiments. We conclude our paper in Section~\ref{sec:future-work}, where we discuss the potential directions of future research.

\section{Related Work}
\label{sec:related-work}

\subsection{Financial Sentiment Analysis}

The first decade of the 21st century marked the advent of the Information age driven by the adoption of Internet as the information exchange medium. Increasingly research has been directed towards tracking financial and economic developments in real time through automated methods utilizing news flow and machine learning techniques. \citeA{wysocki1998cheap} was the first to explore the relation between daily stock message posting volumes, and future stock returns and trading volumes. His observations on 3,000 stocks listed on Yahoo! message boards indicated that the over-night message-posting volumes affect the next day abnormal stock returns and trading volumes. 
	
Consequently, computational linguistics techniques were applied to opinions on various stocks discussed on the stock message boards - Yahoo! Finance \cite{antweiler2004all, das2001yahoo} and RagingBull.com  \cite{antweiler2004all} to identify whether the messages could be translated to financial sentiments - buy, sell, or hold decisions. \cite{das2001yahoo} were among the first to utilize Natural Language Processing (NLP) techniques and human annotated datasets along with various machine learning based classification techniques to automate the information extraction process. Their work also revealed the significance of negation effects on the overall sentiment of the sentence, and the ambiguity inherent in human classification of financial news. Based on the analysis of Morgan Stanley Technology (MSH) index, they concluded that there exists a strong link between aggregate sentiments and market movements. \citeA{antweiler2004all} found evidence that stock messages helped predict market volatility based on a bullishness index derived from more than 1.5 million stock messages about 45 companies in Dow Jones Industrial Average and Dow Jones Internet Index. 
	
While the previous research focused on opinions on stock message boards, \citeA{tetlock2007giving} hypothesized that news columns from major business media firms might contain information that can have significant impact over future company returns and earnings. Based on news data from Wall Street Journal and Dow Jones News Service from 1980 to 2004, and the word polarities (positive, negative and neutral) defined in the General Inquirer (GI) \cite{stone1966general}, he concluded that quantification of financial news presents novel information that can be utilized along with traditional financial information to predict company earnings and stock returns suggesting that the automated news analysis should provide valuable information for traders and analysts. The academic and industrial interest in analyzing news flow and financial text increased after the Global Financial Crisis of 2007-08, which led to the development of the specialized field of fine-grained or aspect-based FSA.

As the sentiment analysis field grew researchers identified that the sentiments are often expressed towards certain aspects, topics or entities which led to the development of the area of aspect-based sentiment analysis or fine-grained sentiment analysis \cite{schouten2015survey}. The aspect-based sentiment analysis (ABSA) corresponds to two tasks - entity (aspect or topic) recognition and subsequently, sentiment extraction for the entity. The entity recognition task led to the development of Named Entity Recognition (NER) systems which are typically conditional random fields models which learn the boundary spans of named entities such as locations, organizations, and companies from given training data \cite{lafferty2001conditional}. Currently, the state-of-the-art publicly available NER systems are trained on 3 datasets - the CoNLL 2003 \cite{sang2003introduction}, MUC-6 \cite{grishman1996design} and MUC-7 \cite{chinchor1997muc}, however, there are no specialized NER systems publicly available for use in finance. The problem of aspect based sentiment analysis has often been formulated as "given an entity, extract the sentiment corresponding to the entity" leading to first SemEval 2016 Task 5 on ABSA which introduced 39 datasets across 8 languages \cite{pontiki2016semeval}. In finance, SemEval 2017 Task 5 subtask 2 and FiQA 2018 Task 1 corresponded to the fine-grained FSA on financial news with training datasets of 1647 and 1750 news texts annotated for entities and their sentiment scores. The FSA research community has focused on the sentiment extraction task, leading to developments in financial lexicon and annotated corpora, and machine learning based approaches for sentiment extraction. We provide a review of the two aspects of fine-grained FSA below.

\subsection{Financial Lexicon and Datasets}

Lexicon-based methods for sentiment analysis utilize sentiment lexicon which contain expressions annotated for their sentiments. These expressions may range from words called opinion words \cite{ding2008holistic} to polarity expressions which can be both subjective and objective expressions \cite{van2015fine}. Generally, the lexicon-based methods extract the sentiment from the text using the semantic orientation of the constituting words or expressions \cite{taboada2011lexicon}. The lexicon-based methods utilize dictionaries of these opinion words or polarity expressions annotated for their sentiment scores or semantic orientations (see also \cite{tong2001operational, pang2008opinion}). There exist multiple sentiment lexicon for general sentiment analysis - Harvard General Inquirer \cite{stone1966general}, MPQA \cite{wiebe2005annotating}, and SentiWordNet \cite{esuli2006sentiwordnet, baccianella2010sentiwordnet}.

\citeA{loughran2011liability}, were the first to study the effects of domain-dependence on prior polarities of words, especially in the financial domain in which the words have interpretations that differ from their general usage. After a detailed study of the Harvard Dictionary, they provide a finance-focused word dictionary with 6 classifications (positive, negative, uncertain, litigious, strong modal and weak modal). \citeA{wilson2005recognizing} noted that the contextual polarities of words in a sentence may not be the same as their prior polarities. They proposed various word-level, sentence-level and document-level features along with machine learning techniques to perform a phrase-level sentiment analysis.

Building on this research, \citeA{malo2013learning}, identified that representation of financial concepts as direction-dependent phrases (ex. profit as ``Positive-if-up", loss as ``Negative-if-up" etc.) is essential to capture the sentiment expressed in the financial text.  To illustrate, the financial concept - ``profit" expresses neutral sentiment in \textit{``The profit stands at \$ 100 million"}, and a positive sentiment in \textit{``The profit increased by 10\%"}. The directional word ``increased" when combined with the word ``profit" makes the sentiment positive in the second sentence. Recognizing that there are multiple ways in which sentiments are expressed in financial text, \citeA{van2015good} developed an annotation scheme to differentiate between two sentiment-bearing expressions - the private state expressions which are explicit expressions (ex. \textit{dissappointing performance}) and the polarity fact expressions which are objective expressions that express sentiment (ex. \textit{facing action from US federal}). Based on their annotation scheme, they provide a bi-lingual annotated corpus of financial text with 4,790 sentences annotated for sentiment bearing expressions and their sentiments and corresponding target entities.

\citeA{balahur2013sentiment} released the JRC Corpus consisting of 1,592 quotes from newspaper articles in English for opinion mining. \citeA{malo2014good} released a collection of around 5000 phrases/sentences manually annotated by 16 subject experts. These were further organized into sets of news headlines with varying inter-annotator agreement. However, these sentences are not annotated for the entity name, making them unusable for entity-aware sentiment extraction. Two other publicly available datasets are from the SemEval 2017 Task 5 Subtask 2 \cite{cortis2017semeval} and FiQA Task 1 Sentiment Scoring Dataset \cite{maia201818} both of which provide a set of 1,000-2,000 news headlines annotated for multiple entities and their sentiment scores ranging between -1 and 1. SemEval 2017 Task 5 Subtask 2 and FiQA Sentiment Scoring dataset offes annotations for multiple entities and sentiment scores including conflicting sentiments, however, the dataset sizes are small and contain limited set of entities. \citeA{van2015fine} provide an annotated corpus of company-specific news for 4 companies, annotating each sentence for its polarity. Their focus however remains on the suitability of their annotation scheme to the financial context. Though they consider news headlines with multiple companies, they provide the annotations of sentiment-bearing expressions and identify the relevant entities. In SEntFiN, we provide sentiment annotations towards the entities in the news headlines and we do not annotate for sentiment-bearing expressions.  

\subsection{Approaches for sentiment extraction}
Research in financial news analysis, especially in the sentiment analysis, started in the 2000's with increasing interest in quantifying and learning the mechanisms through which investor sentiments affect stock movements. Traditional approaches to modelling information from news flow include the count of news items in a certain duration \cite{tetlock2007giving, goonatilake2007volatility}, weighted term vector models \cite{schumaker2008evaluating, schumaker2009textual, wang2012novel}, word-based and sentence-based sentiment analysis \cite{li2014effect} and syntax analysis through recognition of actors and their roles \cite{ding2014using, ding2015deep}. Sentiment analysis-based models have shown to perform better than news count and weighted term vector models in explaining and predicting market movements \cite{li2014news}. Over time, lexicon-based methods and phrase-level sentiment analysis \cite{bruce1999recognizing, riloff2003learning, hu2004mining} which assume that the words have prior and contextual polarities that can be quantified have shown to exhibit better performance than word-based approaches. 

\citeA{malo2014good} study the problem of learning context through the use of lexicon and propose the use of a Linearized Phrase Structure (LPS) model to learn the syntactic structure which performed better than previous baselines.\citeA{van2015fine} are among the first ones to consider the use of annotation tools to identify the polarity bearing expressions, which they term as fine-grained sentiment analysis and show that it performs better than sentence-level sentiment analysis. However, these methods do not address the task of extracting entity-relevant sentiments in the presence of multiple entities. \citeA{pivovarova2018benchmarks} study the problem of extracting entity relevant sentiments, however, their model fails to perform well in the presence of multiple entities with conflicting sentiments. Our approach of tagging entities based on relevance, is able to achieve high level of accuracy in extracting entity-relevant sentiments.

As classification models, SVMs were predominantly used in research \cite{das2001yahoo, antweiler2004all, ranco2015effects, li2014news} until the SemEval 2017 Task 5 Subtask 2 - the fine-grained sentiment analysis of news headlines, in which some of the best performing systems utilized RNNs and LSTMs \cite{mansar2017fortia, moore2017lancaster, ghosal2017iitp, cabanski2017hhu}. With introduction of BERT \cite{devlin2018bert}, we see an uptick in use of language models for sentiment analysis \cite{xu2019bert, sun2019utilizing, gao2019target}.
%\textit{BERT, finBERT, distilBERT, RoBERTa related}

\section{Datasets}
\label{sec:dataset}
In this section, we provide an overview of the two key datasets that we release with this paper. The first dataset that we discuss is our human-annotated financial news dataset with over 14,000 sentiment-entity annotations. We provide a comprehensive overview of the annotation process, discuss the inter-annotator agreement, and provide the dataset statistics. The second dataset is the entity database covering over 1,000 entities and their news representations. The entity database is particularly useful in our work, as our final goal is to evaluate economic value of sentiments for which we are required to map sentiments to entities and construct sentiment indices.

\subsection{SEntFiN 1.0}
\subsubsection{Annotation Process}
In total, 10,753 headlines were sampled from all the financial and economic news pertaining to the duration 2011-2015 from Indian business news provider The Economics Times\footnote{https://economictimes.indiatimes.com/}. The annotation scheme and guidelines were prepared by one of the primary authors, who also acted as the curator of the annotation process. For the annotation process, we identified three human annotators - students pursuing a master's degree in business administration and management and with a strong academic performance in courses relevant to financial markets and macroeconomics. These three human annotators independently approached the curator with their interest in the annotation process, and the curator ensured that the annotators were not informed of the presence of the other annotators until the preparation of the final dataset. Each of the three annotators were provided with the 10,753 news headlines with the entities and their entity spans identified in each news headline by the curator (and validated by the other authors). The annotators were asked to assign a sentiment class - positive, negative, or neutral to the entities identified in the news headlines. For the annotation process, the news headlines were provided to the annotators as a \textit{csv} file with two columns - ``news headline'',``entity-sentiment'' (ex. \{``news headline'': \textit{Negative on Chambal, Advanta: Mitesh Thacker}, ``entity-sentiment'': \{`Chambal': ` ', `Advanta': ` '\} \}). 

Further, the following guidelines were issued to the annotators prior to the annotation process:
\begin{itemize}
    \item The annotators were asked to think like investors and avoid any speculation based on their prior knowledge. The financial sentiments were to be derived from the information that was explicitly available in the headline,
    \item The classes were to be chosen from among one of the three sentiments defined below:
    \begin{enumerate}
        \item \emph{Positive}: A positive sentiment is expressed by the phrase or sub-phrase in relevance to the recognized entity
        \item \emph{Negative}: A negative sentiment is expressed by the phrase or sub-phrase in relevance to the recognized entity
        \item \emph{Neutral}: A neutral sentiment or no sentiment is expressed by the phrase or sub-phrase in relevance to recognized entity or the entity refers to a regulatory organization, a government body etc. for which the concept of a financial sentiment does not apply
    \end{enumerate}
\end{itemize}

Each annotator was provided with a sample of 35 news headlines annotated for the entities and sentiments by one of the primary authors, who also acted as the curator of the annotation process. The annotation process was held over a span of two months, with intermittent interventions by the curator to track the progress and identify potential errors. After the completion of the annotation process by individual annotators, the final dataset preparation process proceeded based on a consensus-driven mechanism as we intended to prepare a single gold-standard. The final decision on the sentiment annotation was driven by a consensus between the three annotators and in the situation where a consensus was not achieved, the final decision on the sentiment annotation was provided by the curator. However, such cases were less than 2\% of the final annotations.  

Overall, each news headline was processed for entities and relevant financial sentiments by three human annotators. Each headline was therefore annotated for the sentiments for identified entities, amounting to a total of 14,404 entities and sentiment annotations. 

\subsubsection{Inter-annotator agreement}
Ambiguity is inherent in human decision making, and especially arises when the same dataset is being annotated by multiple researchers. In order to reduce ambiguity during the annotation process, a sample of 100 annotated headlines, and list of 15 typical ambiguous cases with the rationale for the final recommendation was provided to the annotators as a part of training. The cues for the sources and reasons for ambiguity were derived from the observations in \cite{malo2014good}. After each news headline was annotated by each annotator, and prior to the consensus and final decision making process, the average inter-annotator agreements for each pair of classes were calculated. We observed that there is a high agreement when differentiating negative from positive sentiments (98.26\%) and negative from neutral sentiments (96.85\%). However, as observed in previous studies, there is a lower agreement when differentiating neutral from positive sentiments (80.36\%).

\subsubsection{Dataset}
The output of the annotation process is the SEntFiN 1.0: Sentiment and Entity annotated Financial News dataset. The dataset contains 10,753 news headlines annotated for the entities and their relevant financial sentiments. Of these, 2,847 news headlines have more than two entities among which 1,233 news headlines contain conflicting sentiments. The news headlines with multiple entities contribute $\sim$ 6,500 entity-sentiment annotations, amounting to an average of $\sim$ 2.3 entities per headline. The average word length of sentences in the entire dataset is 9.91 words and sentences with multiple entities tend to have a higher average word length of 10.39 words. SEntFiN identifies 14,404 entities and their sentiments of which positive, negative and neutral sentiments are 35.23\%, 26.48\% and 38.29\% respectively indicating low class imbalance as shown in Table~\ref{tab:sentfin-summary}.

% \usepackage{multirow}
% \usepackage{graphicx}
\begin{table*}[h]
\centering
\caption{\label{tab:sentfin-summary}
Summary statistics of the SEntFiN 1.0 dataset. The dataset provides 14,404 entity-sentiment annotations in 10,753 news headlines. The distribution of entity-sentiment annotations indicates low class imbalance. The 2,847 news headlines with multiple entities contain 6,498 entities indicating an average of 2.28 entities for each news headline.}
\resizebox{\textwidth}{!}{%
\begin{tabular}{c|c|c|cccc}
 & \multirow{2}{*}{\textbf{Headlines}} & \multirow{2}{*}{\textbf{\begin{tabular}[c]{@{}c@{}}Average\\ Sentence Length\end{tabular}}} & \multicolumn{4}{c|}{\textbf{Entity-Sentiment Annotations}} \\ \cline{4-7} 
 &  &  & \textbf{Positive} & \textbf{Negative} & \textbf{Neutral} & \textbf{Overall} \\ \hline
Single Entity & 7,906 & 9.74 & 2,837 (35.88\%) & 2,376 (30.05\%) & 2,693 (34.06\%) & 7,906 \\ \hline
Multiple Entity & 2,847 & 10.39 & 2,237 (34.43\%) & 1,438 (22.13\%) & 2,823 (43.44\%) & 6,498 \\ \hline
\textbf{Overall} & \textbf{10,753} & 9.91 & 5,074 (35.23\%) & 3,814 (26.48\%) & 5,516 (38.29\%) & \textbf{14,404} \\ \hline
\end{tabular}%
}
\end{table*}

\subsection{Entity Database}
\label{sec:entitydatabase}
During the preparation of the dataset for the annotation process, it was observed that the representations of the financial concepts (ex. ``stock" appears as ``stk", ``stocks", ``shares"), and representations of a same company, sector or industry varied with headlines (ex. automotive industry appears as ``auto", ``auto ind", ``auto space"; Larsen {\&} Toubro Ltd. appears ``L{\&}T", ``Larsen", ``Larsen {\&} Toubro", ``Larsen and Toubro"). Traditionally, financial entities corresponding to sectors and industries such as ``pharmaceutical sector", ``automotive sector", ``steel industry" are not recognized as named entities by the NER systems. However, news headlines often express sentiments towards sectors or industries. The performance of the Stanford CoreNLP NER on a set of 250,000 news headlines showed multiple errors as indicated in Table~\ref{tab:ner-results-errors}. 

% \usepackage{multirow}
% \usepackage{graphicx}
\begin{table*}[tbh]
\caption{Major categories of errors identified in the Named Entity Recognition process on 250,000 news headlines using Stanford CoreNLP NER. The errors majorly correspond to wrong entity spans, multiple entities being recognized as single entities, persons recognized as organizations, and missed entities.}
\label{tab:ner-results-errors}
\resizebox{\textwidth}{!}{%
\begin{tabular}{l|l|l}
\hline
\textbf{News   Headline} & \textbf{Entity Recognized} & \textbf{Error} \\ \hline
{\parbox{10cm}{Abbott India Q3 net up 26 pc at Rs 85.56 crore}} & Abbott India Q3 & \multirow{2}{*}{{\parbox{5cm}{Q3 and Q1 indicate   quarters - leading to wrong entity span }}} \\
{\parbox{10cm}{Abbott India Q1 net up 71\% to Rs 51 crore }}& Abbott India Q1 &  \\ \hline
{\parbox{10cm}{ Hold Abbott India Ltd., target Rs 4480.0 : Centrum Broking}} & Hold Abbott India Ltd. & \multirow{2}{*}{{\parbox{5cm}{sentiment bearing expressions wrongly recognized as entities}} } \\
{\parbox{10cm}{ Buy Apollo Hospitals Ltd with a target of Rs 1366: Sandeep Wagle }}& Buy Apollo Hospitals Ltd &  \\ \hline
{\parbox{10cm}{ Accumulate Tata Steel, Hindalco on   declines: Ashit Suri }} & Accumulate Tata Steel & \multirow{2}{*}{{\parbox{5cm}{wrong entity span as   well as missing entities }}} \\
{\parbox{10cm}{ Accumulate Tata Steel, Shree Cement stocks: JV Capital Services }} & Accumulate Tata Steel &  \\ \hline
{\parbox{10cm}{ Indian IT to battle IBM, Accenture, Hewitt over \$12 bn deals }} & Accenture, Hewitt & \multirow{5}{*}{{\parbox{5cm}{Multiple entities recognized as a single entity; in a few cases sentiment bearing expressions   also identified as entity }}} \\
{\parbox{10cm}{ Accumulate Coal India \& Bharti Airtel on correction: Sajiv Dhawan }} & Accumulate Coal India \&   Bharti Airtel &  \\
{\parbox{10cm}{ After Airtel \& Idea, Vodafone announces prepaid free calling plans }} & Airtel \& Idea &  \\
{\parbox{10cm}{ Spectrum Auction: Bharti Airtel \& Idea Cellular outperforms broader market }}& Bharti Airtel \& Idea Cellular &  \\
{\parbox{10cm}{ CAG reports on Air India \& RIL   trigger fears of policy paralysis }} & Air India \& RIL &  \\ \hline
{\parbox{10cm}{ Anand Mahindra \& Rajiv Bajaj chasing   growth in automobile sector with contrasting strategies }}& Anand Mahindra \& Rajiv Bajaj & {\parbox{5cm}{persons recognized as   organizations, multiple entities recognized as a single entity }} \\ \hline
{\parbox{10cm}{ Corporation, Andhra \& Dena Bank hot   picks on M\&A street }} & Andhra \& Dena Bank & {\parbox{5cm}{ multiple entities recognized as   a single entity }}\\ \hline
\rule{0pt}{0.65cm}{\parbox{10cm}{ Technologies like ArtificiaI   Intelligence, big data impacting power sector: Tata Power }} & ArtificiaI Intelligence & {\parbox{5cm}{wrong entity recognition }} \\ \hline
\end{tabular}%
}
\end{table*}

To effectively solve the problem of entity recognition and sentiment attribution, we realized that a database that provides information on such representations are necessary, as they are not available as public resources. We therefore manually studied 920 companies which were included in the NSE500 broad-based market index (represents over 95\% free float market capitalization in the Indian equity market) in the duration 2002 to 2017 for which the news headlines were available, and 89 other entities (sectors, commodities, and currencies) that appeared in the news headlines (appearing in the SEntFiN dataset) provided by The Economic Times\footnote{https://economictimes.indiatimes.com/} and Moneycontrol\footnote{https://www.moneycontrol.com/} and identified their various media representations amounting to 5,070 phrases. Due to our focus on the companies which were publicly traded and were a part of the NSE500 index the entity database is not an exhaustive list of companies, however, it provides a representative list of the equity market in India for the chosen duration.

Each database entry contains three values - first, the symbol assigned to the entity (stock ticker for a company); second - the official name of the entity if present; and third - the list of all phrases that refer to the entity in news media. For example, the company \emph{State Bank of India} has the following database entry: \{``SBIN" - \{\emph{Official Name}: State Bank of India Ltd.\}, \{\emph{Other forms}: State Bank, SBI, State Bank of India\} \}. We make this database publicly available along with this work. In our future work, we aim to include more entities to the database. The various representations of financial concepts such as ``stk", ``stks" etc. were added to the financial lexicon used in the study.

\section{Sentiment Extraction}
\label{sec:learning-schemes}
In this section, we describe our choices of learning schemes which is a combination of sentence representations and classifier methods, and the hyper-parameter choices utilized in our experiments. Further, we present and interpret the results of test performance of all learning schemes.

\subsection{Sentence Representations}
Each instance of the headline generated after the recognition of entities and removal of noise are then transformed into input feature vectors based on their sentence representations. We categorize our choice of the sentence representations based primarily on the approach taken to extract features which is either through financial lexicon or through pre-trained representations. We discuss these further below.

\subsubsection{Lexicon-based representations}

We use MPQA \cite{wiebe2005annotating}, GI \cite{stone1966general} and LM \cite{loughran2011liability} dictionaries to capture the prior polarities. If an overlap is encountered, the prior polarity as defined by LM dictionary is preferred over other dictionaries. We utilize the financial concepts and directionality words list developed by \cite{malo2013learning}. Contextual valence shifters (negators) are derived from GI. Numbers are identified and represented by ``Number" feature.

We note that lemmatization of words typically results in loss of features majorly pertaining to the prior polarities and directional dependencies. For example, in ``Infosys' stock rallies", ``rallies" is used to express upward movement in the ``Infosys' stock price" indicating a \emph{positive} sentiment towards ``Infosys", while in ``Nifty rally ends", ``rally" describes an action already in place and sentiment for ``Nifty" is \emph{neutral}. 

After an investigation of the SEntFiN 1.0, a custom dictionary containing 970 words commonly used in the financial news were prepared. Of the 970 words, 895 words are present in the LM dictionary with differing annotations, and the remaining 75 words do not appear in any dictionary. For the purpose of feature extraction, we define Lexicon as including  LM, MPQA, GI, Malo et al, and our custom dictionary as shown in Table~\ref{tab:lexicon}. The feature annotations based on the custom dictionary\footnote{Available upon request from the corresponding author} are given preference over the LM dictionary during the feature extraction step.

\begin{table*}[htb]
\caption{Lexicon has 8 types of word-level features and annotations for 94,965 words. In case of an overlap, the preference order is as follows: direction-dependency $>$ directionality $>$ negation $>$ prior sentiment}
\label{tab:lexicon}
\resizebox{\textwidth}{!}{%
\begin{tabular}{lllllll}
\hline
\textbf{Features} & \textbf{LM} & \textbf{MPQA} & \textbf{GI} & \textbf{Malo et al.} & \textbf{Custom} & \textbf{Overall} \\ \hline
positive & 408 & 2,718 & 55 & - & 276 & 3,457 \\
neutral & 82,319 & 572 & 144 & - & 21 & 83,056 \\
negative & 2,404 & 4,911 & 82 & - & 85 & 7,482 \\
Up & 4 & - & 128 & - & 233 & 365 \\
Down & - & - & 122 & - & 213 & 335 \\
Positive-if-up & - & - & - & 69 & 100 & 169 \\
Negative-if-up & - & - & - & 28 & 32 & 60 \\
negator & - & - & 31 & - & 10 & 41 \\ \hline
\multicolumn{7}{l}{Lexicon = LM + MPQA + GI + Malo et al. + Custom (our additions and modifications)} \\ \hline
\end{tabular}%
}
\end{table*}

% \begin{itshape}
% Let F = \{ $f_{1}$, ..., $f_{N}$ \} denote the collection of N word-level features, and let S be the space of possible sentences. The \textbf{word-level feature extraction step} can be defined as a transformation w : s \mapsto \mbox{[} \, $w_{1}$, $w_{2}$, . . . , $w_{k}$\mbox{]} \, ,  which presents the sentence as a vector of k word-level features, $w_{i}$ \in F, i \in \mbox{\{}1, . . . , k\mbox{\}}
% \end{itshape}

\begin{figure*}[h]
\centering
	\includegraphics[width=\textwidth,keepaspectratio]{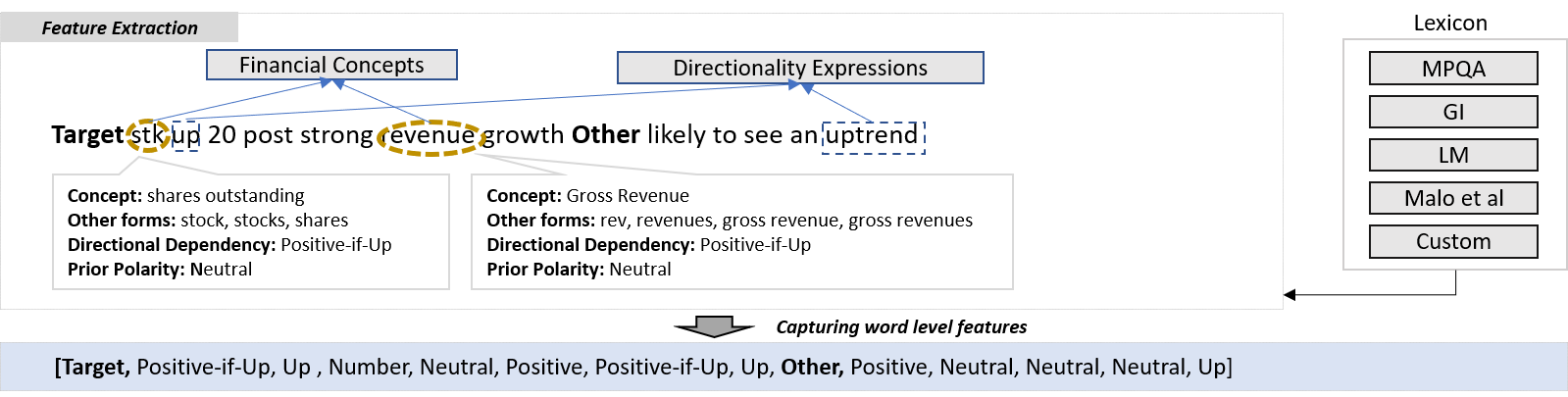}
	\caption{The lexicon-based feature extraction step utilizes lexicon to generate a vector of features corresponding to the tokens in the sentence. We do not employ stop-word removal in our pre-processing step.}
	\label{fig:fe_ex}
\end{figure*}

Building upon the example in Figure~\ref{fig:me_handle}, we depict the lexicon-based feature extraction for the first instance in Figure~\ref{fig:fe_ex}. We do not employ stop-word removal, as the number of tokens in the entire dataset is not large to ignore the frequencies of stop-words. The vector of features is then modeled based on one of the two approaches mentioned below.

\emph{\textbf{Linearized Phrase Structure}} (LPS) models were introduced by \cite{malo2014good} as a simple and effective technique to capture syntactic structure of news headlines. The LPS model, captures the syntactic structure of the sentence utilizing a sequence of literals framework, where in the literals are the word-features. The final sequence is then converted into a bit vector using a binary coding scheme. 
% With the new features \textbf{Target} and \textbf{Other}, the LPS model will be able to capture the various entities and their relevant contexts from the news headline.

\emph{\textbf{UBT}} N-gram based models have been among the first techniques to be utilized to capture the probability of co-occurrence of words in text and have been shown to perform reliably well in text classification tasks \cite{cavnar1994n}. We note that, n-grams in the context of sentiment extraction from news headlines have the ability to capture the syntactic structure and we apply the n-gram model after extracting word-level features, rather than on words itself. We define UBT as the Uni-gram + Bi-gram + Tri-gram ensemble vector of term frequencies based on word-features.

% \begin{itshape}
% 	Let F be as defined above, and consider a transformation w(s) for a specific phrase s. We define 3 vectors $V_{1}$, $V_{2}$, $V_{3}$  such that,
% 	\[
% 	V_{1} = [n_{f_{1}}, n_{f_{2}},..., n_{f_{N}}] \] \[
% 	V_{2} = [n_{( \, f_{1}, f_{1} )\,}, n_{( \, f_{1}, f_{2} )\,},...,n_{( \, f_{N}, f_{N} )\,} ] \] \[
% 	V_{3} = [ n_{( \, f_{1}, f_{1}, f_{1} )\,}, n_{( \, f_{1}, f_{1}, f_{2} )\,},...,n_{( \, f_{N}, f_{N}, f_{N} )\,}] \]
% 	where n denotes the number of occurrences of the tuples corresponding to the co-occurrences of word-level features in the mapping f(s). \\
% % 		We further define the various phrase-structure models based on the n-grams as follows:
% 	%The Uni-gram (U) model by U = [$V_{1}$] \\
% 	%The Uni-gram + Bi-gram (UB) model by UB = [$V_{1}$, $V_{2}$] \\
% 	%The Uni-gram + Bi-gram + Tri-gram (UBT) model  by UBT = [$V_{1}$, $V_{2}$, $V_{3}$]\\
% 	%The Uni-gram + Bi-gram + ... + N-gram (N) model by N = [$V_{1}$, $V_{2}$, $V_{3}$, ..., $V_{n}$]
% \end{itshape}
% We further define the Uni-gram + Bi-gram + Tri-gram (\textbf{UBT}) model  by \textbf{UBT = [$V_{1}$, $V_{2}$, $V_{3}$]}.

\subsubsection{Pre-trained representations}
\label{sssec:pre-trained}
For this study, we have used both context-free representations and context-aware representations, elaborated further below.

\textbf{Context-independent representation}: We use GloVe~\cite{pennington2014glove} for context-independent representation of words. For our study, we use the GloVe word-embeddings dataset with 840 billion tokens, 2.2 million vocabulary and each word/token is represented as vector of 300 dimensions. GloVe embeddings are derived based on the optimization of the word embedding vectors in the word analogy task. 

\textbf{Context-aware representation:} Bi-directional Encoder Representations from Transformers (BERT) is a deep bi-directional general language model consisting of stacked layers of a set of transformer encoders. BERT uses either 12 or 16 layers of transformers which generates a context-aware representation for words. The training data used for BERT is 16 GB corpus of books and Wikipedia. For our study, we utilize finBert~\cite{araci2019finbert}, a financial domain focused BERT implementation which is fine-tuned using Financial PhraseBank~\cite{malo2014good} and FiQA Task 1 sentiment scoring dataset. Further, we compare two other versions of BERT - RoBERTa \cite{liu2019roberta} which is an optimized BERT and trained over a 160 GB corpus (144 GB additional corpus than BERT) and DistilBERT \cite{sanh2019distilbert} which has less model parameters and therefore a smaller version of BERT. We use the source code of finBERT released by ~\cite{araci2019finbert}. For RoBERTa and DistilBERT, we implement models publicly made available on HuggingFace \cite{wolf2019huggingface}, using \textit{roberta-base-openai-detector} and \textit{distilbert-base-uncased} implementations respectively \footnote{\url{https://www.huggingface.co/transformers/pretrained\_models.html}}.
%\footnote{\url{https://huggingface.co/transformers/pretrained_models.html}}.

\subsection{Classification Algorithms}\label{subsec:algo}
\emph{\textbf{Support Vector Machines}} (SVMs) are a family of kernel-based binary classification methods originally introduced by Cortes and Vapnik \cite{cortes1995support, vapnik2013nature}, which maximize the margin defined by the distance between the nearest points (support vectors) to a hyper-plane that separates the two classes in the feature space. We implement SVM with a linear kernel using the scikit-learn package~\cite{scikit-learn}.

\emph{\textbf{Gradient Boosting Machines}} (GBMs) are a family of ensemble boosting classifier methods, which formulate the classification problem as a loss minimization problem, and approach the solution using gradient descent approach \cite{friedman2001greedy}. The base learners of GBMs are typically Classification and Regression Trees (CARTs) \cite{breiman2017classification} which recursively partition the feature space and fit simple regression/classification models to each partition. This tree type model is implemented with XgBoost package~\cite{chen2016xgboost}.

\emph{\textbf{Multi-layer Perceptrons}} (MLPs) are feed-forward Artificial Neural Network. Each node in an MLP acts as a non-linear activation function, except in case of nodes in input layer. In this study we implement MLP with scikit-learn package~\cite{scikit-learn}.

\emph{\textbf{Recurrent Neural Networks}} (RNNs) are successive layers of artificial neurons proposed by~\citeA{rumelhart1986learning}. RNNs build over MLPs and add a hidden state which is the output of the previous time step, therefore, are suitable for data with sequential dependencies such as text.

\emph{\textbf{Long-Short Term Memory}} (LSTM)~\cite{hochreiter1997long} is a version of RNN which aims to learn long-term dependencies of a given sequence and solve the vanishing gradient problem of vanilla RNNs. Bi-directional LSTM (BiLSTMs)~\cite{graves2005framewise} are similar to BiRNNs in their construction.

We construct uni/bi-directional RNN and LSTM model using the Keras library in Python ~\cite{chollet2018keras} with Tensorflow backend~\cite{abadi2016tensorflow}.

\subsection{Experiments}
\label{sec:experiments}
As a first step, we processed the headlines in SEntFiN datasets to handle multiple entities using the features \textbf{Target} and \textbf{Other}. The processed dataset contains 14,404 headlines with relevant entities identified as \textbf{Target} with their corresponding sentiment label, and the irrelevant entities identified as \textbf{Other} if present. 

For experiments with the lexicon-based and GloVe-based approaches, we divided the processed dataset using a 80/20 train/test split, leading to $\sim$11,200 entity-sentiment annotations in the training set and $\sim$2,800 entity-sentiment annotations in the testing set. We train our learning schemes on a training set, compare the performance on a test set, and evaluate based on Accuracy and F1-score for each sentiment class separately. Given that the distribution of news headlines in the dataset is not uniform with respect to headlines with multiple entities and single entities, we chose to create 31 different randomly sampled train/test datasets instead of a k-fold cross-validation approach. We report the results as the median performance across the 31 different train/test splits in the following section. The average and standard deviation of the results are reported in Appendix~\ref{sec:mean-sd-ml-experiments}. For the experiments with the BERT-based models, given that the language models are large with over millions of parameters, we utilized a 5-fold cross validation approach with a 80-10-10 train-validation-test split.

% \subsection*{Implementation Details}
% \label{subsec:implement}
% As the first step, we pre-process the SentiFiN dataset to replace the relevant companies name with Entity\_0 and Entity\_1 based on the relevance of the corresponding sentiment label. A company for which the sentiment is present in the sentiment label is has its name replaced with Entity\_1 and all other companies, if any, present in the same headline is replaced with Entity\_0. This process is repeated to accommodate sentiments for multiple companies, if present, in a single headline. We then shuffle the dataset and do a train-test split of 80\% - 20\% across all three labels for all the models mentioned in the experiment. In case of FinBERT, the training data is further split into 80\% - 20\% in training and validation set.

\textbf{Implementation Details:} We use default parameters for SVC (scikit-learn) and XgBoost Classifier. We constructed the MLP as a 10x10 network with other parameters for the model set to default. The uni/bi-directional layer implementations of LSTM and RNN each have 100 nodes. All four implementations of LSTM and RNN are optimized with \textbf{ADAM}~\cite{kingma2014adam}, with a default learning rate of \textbf{0.01} and uses sigmoid activation function in the output layer with 10 epochs. A dropout probability of \textit{p = 0.2} is applied after the embedding layer and LSTM/RNN layers. 

We use finBERT \cite{araci2019finbert} and further fine-tuned the model using our SEntFiN 1.0 dataset for training. For our implementation of finBERT, we use a warm-up proportion of 0.2, a dropout probability of \textit{p = 0.2}, a learning rate of 2e-5, a mini-batch size of 64 and a maximum sequence length of 30 tokens. For RoBERTa, after a comparison of various hyper parameter choices, we report the results of two models -  RoBERTa (A) with a dropout probability of \textit{p - 0.1}, weight decay of 0.02, and a learning rate of 2e-5, and RoBERTa (B) with a dropout probability of \textit{p = 0.2}, weight decay of 0.01, and a learning rate of 2e-5. For both RoBERTa models, we use a mini-batch size of 64 and a maximum sequence length of 30 tokens. For DistilBERT, after a comparison of various hyper parameter choices, we report the results of two models -  DistilBERT (A) with a dropout probability of \textit{p - 0.1}, weight decay of 0.02, and a learning rate of 2e-5, and DistilBERT (B) with a dropout probability of \textit{p = 0.2}, weight decay of 0.01, and a learning rate of 2e-5. We use a 5-fold approach with 10 epochs each for RoBERTa and DistilBERT to ensure that a wide combination of data for the train-validate-test split of the dataset is considered.

\subsection{Results and Discussion}
\label{sec:results}

In Table \ref{tab:ml-experiments-1}, we present the results of the NLTK Vader and HuggingFace Sentiment Analysis systems on the SEntFiN news headlines dataset. In such kind of non-entity aware systems, a single sentiment is assigned to a headline, and in case of multiple entities within the headline, the same sentiment is assumed for all the entities. Since these systems are not trained for entity-aware sentiment extraction task, the accuracies and F1-scores are clearly quite low. 

%on both the positive and neutral sentiments are much lower than the least performing learning scheme which utilizes LPS and SVM in our entity-sentiment extraction approach.

% \usepackage{graphicx}
\begin{table*}[htb]
\centering
\caption{A comparison of non-entity aware sentiment analysis approaches on the SEntFiN dataset. NLTK Vader utilizes a lexicon-based approach with simple count-based features. For HuggingFace Sentiment system, we utilized the Twitter-roBERTa-base model which is trained on $\sim$58 million tweets and finetuned for sentiment analysis with TweetEval benchmark.  }
\label{tab:ml-experiments-1}
\resizebox{\textwidth}{!}{%
\begin{tabular}{lllllll}
 & \multicolumn{2}{c}{positive} & \multicolumn{2}{c}{negative} & \multicolumn{2}{c}{neutral} \\ \cline{2-7} 
 & Accuracy & F1-Score & Accuracy & F1-Score & Accuracy & F1-Score \\ \hline
\multicolumn{1}{l|}{NLTK Vader} & 65.17\% & \multicolumn{1}{l|}{64.09\%} & 74.02\% & \multicolumn{1}{l|}{52.68\%} & 39.70\% & \multicolumn{1}{l|}{77.61\%} \\
\multicolumn{1}{l|}{HuggingFace Sentiment} & 74.90\% & \multicolumn{1}{l|}{76.30\%} & 82.35\% & \multicolumn{1}{l|}{82.02\%} & 58.79\% & \multicolumn{1}{l|}{60.59\%} \\ \hline
\end{tabular}%
}
\end{table*}

In Table \ref{tab:mlexperiments-2}, we present the results of the experiments in which the models have been trained for the entity-aware sentiment extraction task. In case of the lexicon-based sentence representations, we observe that the ensemble UBT approach of capturing features performs better than the previous established benchmark of LPS approach across both SVM and GBM classification methods with prominent improvements across positive and neutral classes. GBM achieves better classification performance on both the lexicon-based representations with 3-5\% increments in both Accuracy and F1-score. With the use of MLPs we do not achieve increments in performance as expected and we also observe that UBT+GBM learning scheme performs better than the current UBT+MLP learning scheme. It is however possible that with appropriate hyper-parameter configuration, MLPs may perform better than GBMs.

% \usepackage{multirow}
% \usepackage{graphicx}
\begin{table*}[htb]
\caption{Performance summary of the 12 learning schemes. The lexicon-based approaches utilize the custom dictionary consisting of 94,965 word-feature annotations.}
\label{tab:mlexperiments-2}
\resizebox{\textwidth}{!}{%
\begin{tabular}{cccccccc}
\hline
\multicolumn{1}{l}{} & \multicolumn{1}{l}{} & \multicolumn{2}{c}{Positive} & \multicolumn{2}{c}{Negative} & \multicolumn{2}{c}{Neutral} \\ \hline
Approach & Learning Scheme & Accuracy & F1-Score & Accuracy & F1-Score & Accuracy & F1-Score \\ \hline
\multicolumn{1}{c|}{\multirow{5}{*}{Lexicon}} & \multicolumn{1}{c|}{LPS + SVM} & 79.17\% & \multicolumn{1}{c|}{69.58\%} & 85.00\% & \multicolumn{1}{c|}{71.48\%} & 73.69\% & \multicolumn{1}{c|}{66.81\%} \\
\multicolumn{1}{c|}{} & \multicolumn{1}{c|}{LPS + GBM} & 82.45\% & \multicolumn{1}{c|}{74.72\%} & 87.41\% & \multicolumn{1}{c|}{76.45\%} & 77.41\% & \multicolumn{1}{c|}{71.18\%} \\ \cline{2-8} 
\multicolumn{1}{c|}{} & \multicolumn{1}{c|}{UBT + SVM} & 83.79\% & \multicolumn{1}{c|}{77.24\%} & 87.24\% & \multicolumn{1}{c|}{75.85\%} & 79.38\% & \multicolumn{1}{c|}{73.07\%} \\
\multicolumn{1}{c|}{} & \multicolumn{1}{c|}{UBT + GBM} & 85.00\% & \multicolumn{1}{c|}{78.73\%} & 88.48\% & \multicolumn{1}{c|}{77.81\%} & 80.52\% & \multicolumn{1}{c|}{74.91\%} \\
\multicolumn{1}{c|}{} & \multicolumn{1}{c|}{UBT + MLP} & 82.38\% & \multicolumn{1}{c|}{75.58\%} & 86.83\% & \multicolumn{1}{c|}{74.13\%} & 78.24\% & \multicolumn{1}{c|}{72.29\%} \\ \hline
\multicolumn{1}{c|}{\multirow{8}{*}{Pre-trained}} & \multicolumn{1}{c|}{GloVe + RNN} & 83.66\% & \multicolumn{1}{c|}{76.28\%} & 86.49\% & \multicolumn{1}{c|}{74.53\%} & 79.03\% & \multicolumn{1}{c|}{71.71\%} \\
\multicolumn{1}{c|}{} & \multicolumn{1}{c|}{GloVe + Bi-RNN} & 82.99\% & \multicolumn{1}{c|}{75.41\%} & 85.83\% & \multicolumn{1}{c|}{72.88\%} & 78.21\% & \multicolumn{1}{c|}{71.20\%} \\
\multicolumn{1}{c|}{} & \multicolumn{1}{c|}{GloVe + LSTM} & 84.73\% & \multicolumn{1}{c|}{78.19\%} & 87.19\% & \multicolumn{1}{c|}{76.16\%} & 80.60\% & \multicolumn{1}{c|}{73.99\%} \\
\multicolumn{1}{c|}{} & \multicolumn{1}{c|}{GloVe + Bi-LSTM} & 84.78\% & \multicolumn{1}{c|}{78.49\%} & 86.85\% & \multicolumn{1}{c|}{75.69\%} & 80.78\% & \multicolumn{1}{c|}{74.32\%} \\ \cline{2-8} 
\multicolumn{1}{c|}{} & \multicolumn{1}{c|}{finBERT} & 90.58\% & \multicolumn{1}{c|}{92.80\%} & 93.22\% & \multicolumn{1}{c|}{\textbf{95.11\%}} & 89.45\% & \multicolumn{1}{c|}{\textbf{91.90\%}} \\
\multicolumn{1}{c|}{} & \multicolumn{1}{c|}{DistilBERT (A)} & 94.34\% & \multicolumn{1}{c|}{92.20\%} & 94.48\% & \multicolumn{1}{c|}{89.20\%} & 90.89\% & \multicolumn{1}{c|}{88.10\%} \\
\multicolumn{1}{c|}{} & \multicolumn{1}{c|}{DistilBERT (B)} & 94.55\% & \multicolumn{1}{c|}{92.40\%} & 94.34\% & \multicolumn{1}{c|}{89.00\%} & 90.82\% & \multicolumn{1}{c|}{88.00\%} \\
\multicolumn{1}{c|}{} & \multicolumn{1}{c|}{RoBERTa (A)} & 95.24\% & \multicolumn{1}{c|}{93.40\%} & \textbf{95.31\%} & \multicolumn{1}{c|}{91.00\%} & 92.34\% & \multicolumn{1}{c|}{89.90\%} \\
\multicolumn{1}{c|}{} & \multicolumn{1}{c|}{RoBERTa (B)} & \textbf{95.38\%} & \multicolumn{1}{c|}{\textbf{93.60\%}} & 95.10\% & \multicolumn{1}{c|}{90.50\%} & \textbf{92.41\%} & \multicolumn{1}{c|}{90.10\%} \\ \hline
\end{tabular}%
}
\end{table*}

In case of the learning schemes based on pre-trained representations, the GloVe word embeddings with LSTMs perform better than the RNNs with improvements across both accuracy and F1-score, however, the differences are not pronounced. The BERT based learning schemes outperforms all GloVe based learning schemes with average increments of over 7\% in accuracy and 15\% in F1-score compared to the best performing GloVe-based learning scheme.Among the BERT based learning schemes, RoBERTa achieves the highest accuracy scores across all the sentiment classes, with above 95\% accuracy in positive and negative classes, and $\sim$92\% accuracy in neutral class. finBERT achieves the best F1-score across negative and neutral classes, with a marginally lower F1-score compared to RoBERTa-based learning schemes in the positive class.

%\st{RoBERTa reports the best performance across learning schemes with accuracy improvements of 5.58\%, 4.74\% and 8.93\% and F1-score improvements of 14.07\%, 17.3\% and 16.99\% in comparison to the best-performing lexicon-based learning scheme (UBT+GBM) across positive, negative and neutral classes respectively.}

We observe that UBT+GBM learning scheme achieves comparable performance with GloVe based learning schemes, which is indicative of the potential of n-gram models in short text. The ability to achieve the performance of deep-learning approaches through feature-engineering based on specialized domain knowledge is noteworthy. The training process for generating the GloVe word embeddings includes the use of both the local and global context optimized towards achieving best performance on word analogies \cite{manning2014stanford}. As such, these unsupervised models learn from large datasets such as Wikipedia, Twitter, among others where each word appears in multiple contexts with multiple meanings (also known as polysemy). Since, the training process optimizes for analogies across multiple contexts, the embeddings learn a large number of weights. However, we observe that the contextual embeddings fail to perform effectively in a domain specific task as in the case of financial sentiment analysis. Even with large number of weights, the GloVe embeddings likely suffer similar disadvantages of a general sentiment lexicon \cite{loughran2011liability}. The high performance of the UBT models are likely to arise from two key reasons. First, UBT models are effective on short text and in the study we utilized UBT on the features rather than on words likely leading to higher contextual performance. Second, The lexicon used in the study is customized to the financial sentiment analysis task. In addition to changes to the existing lexicon, we have added 75 new words to the financial lexicon. We believe that the lexicon provided good word-feature annotations which the UBT model utilized effectively.

Most of the learning schemes report highest performance on \textit{negative} class, lowest performance on \textit{neutral} class and medium performance on \textit{positive} class. This indicates that the models are better able to differentiate between the negative sentiments from positive or neutral sentiments, which has been observed by \cite{malo2014good} as well. However, RoBERTa based models achieve high accuracy scores on positive and neutral classes unlike other learning schemes. BERT language models utilize a masking strategy in which hidden sections of text are learnt from the datasets, through which the contextual embeddings of tokens are generated. Therefore, fine-tuning BERT to the task-at-hand using the approach as performed in this study provides high performance in a domain-specific task, here financial sentiment analysis. RoBERTa is learnt on much larger dataset than BERT including a novel Common Crawl CC-News dataset \cite{mackenzie2020cc} containing over 44 million news documents. This learning procedure is likely to have produced an improved language model and improved performance in the financial sentiment extraction task.

\section{Information Content of News Flow}
\label{sec:informationContent}

The information related to the value of securities traded publicly, is reflected in the prices set by the forces of supply and demand on the exchanges that the securities trade on. The market-based pricing mechanism constantly consumes the publicly available information related to the historical performance and the expected future performance of the underlying asset. This mechanism was shown to be profound in emerging markets \cite{morck2000information} where the stock price variations were not much affected by fundamentals. With respect to stocks, major public announcements such as earnings releases, changes in management, new product launches are bound to have an impact on stock prices, and have been well studied and documented \cite{lee2009immediate, chambers1984timeliness, warner1988stock}. While, these are the major approaches undertaken by companies to release company related information to the public, there exists an unstructured form of news flow which involves opinions, discussions, daily events regarding companies, which may not necessarily involve company's participation. The news released by media houses fall into this category of news flow, where the source (the editor) determines the content pertaining to the reported market events. Often, such information is utilized by High Frequency Traders in extremely short durations (orders of $10^{-9}$ to $10^{-6}$ seconds) to conduct directed trades. However, it is of interest to identify whether such news flow has economic value over longer time horizons.

\subsection{Modelling information from news flow}
The most prominent approach to modelling information from news flow previously has been to compute a measure for sentiment in a certain duration based on the number of buy and sell decisions derived from the news headlines \cite{das2001yahoo, zhang2010trading}. The sentiment measure is constructed using the following formula,  

The sentiment score (\textit{s}) for a specific entity E, and for a certain duration T, is measured by the following formula,
\[ s^{1}_{T,E} = \frac{Pos_{T,E}{-Neg_{T,E}}}{Pos_{T,E}+Neg_{T,E}} \]
where, $Pos_{T,E}$, $Neu_{T,E}$ are the number of Positive and Negative recommendations

However, this measure does not account for the news flow that contains neutral sentiment. Neutral sentiments in a certain duration are likely to attenuate the effects of positive and negative sentiments therefore leading to reduced movements in market prices. Given this inverse relationship of price movements and neutral sentiments, we construct an alternative sentiment measure as the difference between the frequencies of buy and sell decisions relative to all the decisions (buy, sell and hold) generated with respect to a entity in a certain duration. This construction takes into account the probability that the news source generates neutral sentiments, and attenuates the effects of positive and negative sentiments in a neutral environment.

\[ s^{2}_{T,E} = \frac{Pos_{T,E}{-Neg_{T,E}}}{Pos_{T,E}+Neu_{T,E}+Neg_{T,E}} \]
where, $Pos_{T,E}$, $Neu_{T,E}$, $Neg_{T,E}$ are the number of Positive, Neutral, and Negative recommendations

\subsection{Datasets}
We utilize two datasets for our experiments, one corresponding to the aggregate market movements, and the other corresponding to the information from news flow, as detailed below.
\subsubsection{\textbf{NSE 500 index prices}}
The NSE 500 index is the broad equity market index based on free-float market capitalization tracking more than 95\% of the market capitalization at any point of time\footnote{NSE India official website}. The index has a base value of 1000, established on 01 January, 1995. The index is considered for reconstitution semi-annually, and more than 400 companies have been changed since inception. We collect the daily opening and closing values of index from 01 January, 2002 to 31 December, 2017.
\subsubsection{\textbf{News Flow}}
We captured the news headlines along with the timestamp for the duration 01 January, 2002 to 31 December, 2017 relevant for the NSE 500 index. Overall, we collected 418,703 news headlines relevant to the time duration, which led to 576,191 instances of news text after our entity-recognition step using the entity database\footnote{We cannot release the data publicly due to legal reasons. Available upon request for academic research}. We extracted sentiments for the recognized entities information utilizing the best performing RoBERTa-based model.  

\begin{figure*}[!htbp]
    \includegraphics[width=\textwidth,keepaspectratio]{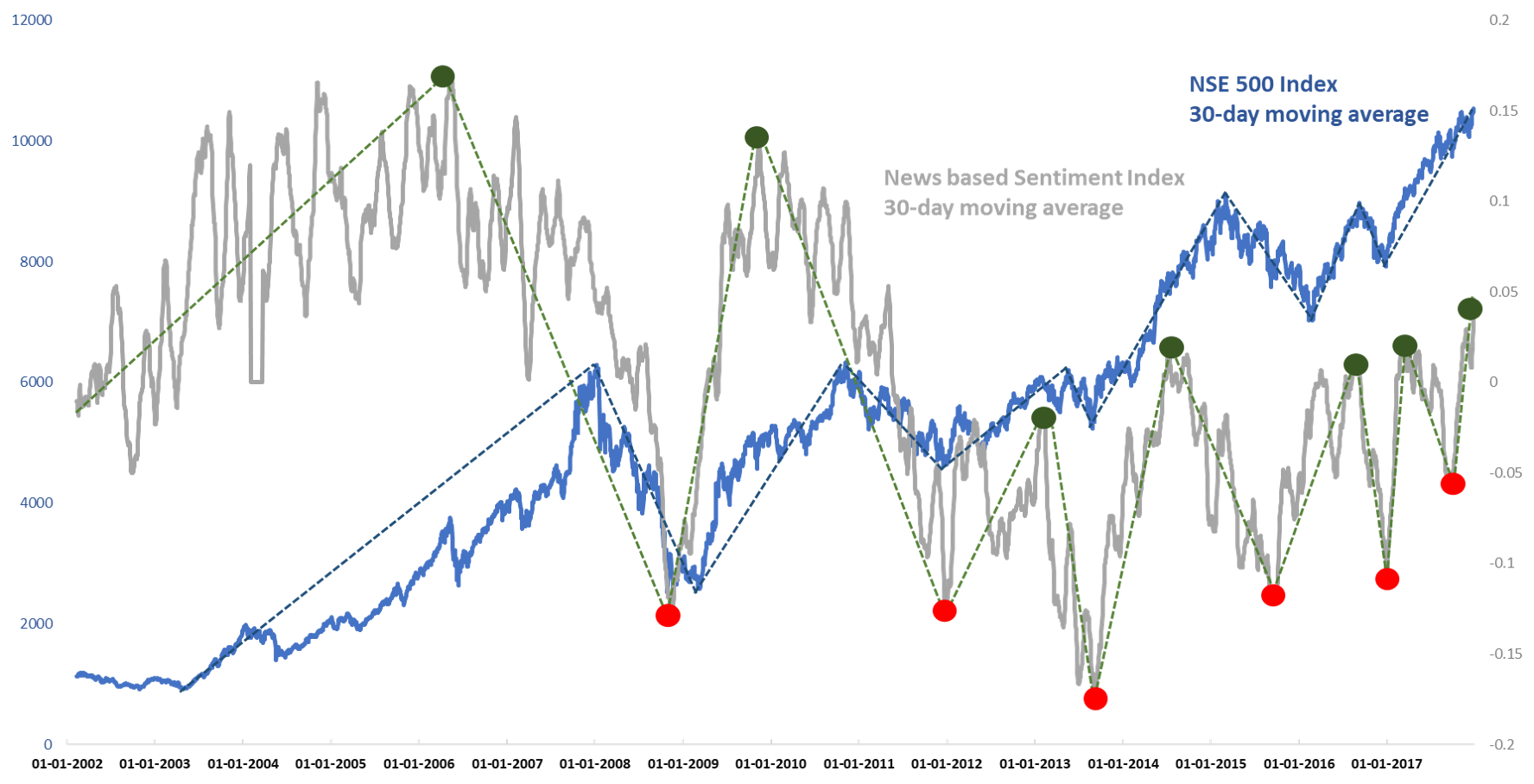}
    \caption{A snapshot of the 30-day moving averages of the NSE 500 and daily News-based Sentiment indices over the duration 01 January 2002 - 29 December 2017. The graph is derived from the data corresponding to 5,113 market days and 576,191 instances of news headlines. The graph indicates that the movements in the news-based sentiment index and the NSE 500 index correlate, with the troughs and crests in sentiment index often appearing earlier than in NSE 500.}
    \label{fig:mkt_movement}
\end{figure*}

In Figure~\ref{fig:mkt_movement}, we juxtapose the 30-day moving averages of the daily news-based sentiment index and the NSE 500 index. We can observe that there is an apparent correlation between the long-term movements in both indices, with sentiments movements preceding the market movements. Moreover, we find that the temporal distance between the movements in sentiments and markets, has reduced over time, likely due the technological progress leading to improvements in market efficiency. Nevertheless, there appears to be relationship between the sentiments and market movements, and therefore, we developed an experiment to evaluate the economic value of sentiments.

\subsection{Construction of the experiment}
\begin{figure*}[!htbp]
    \includegraphics[width=\textwidth,keepaspectratio]{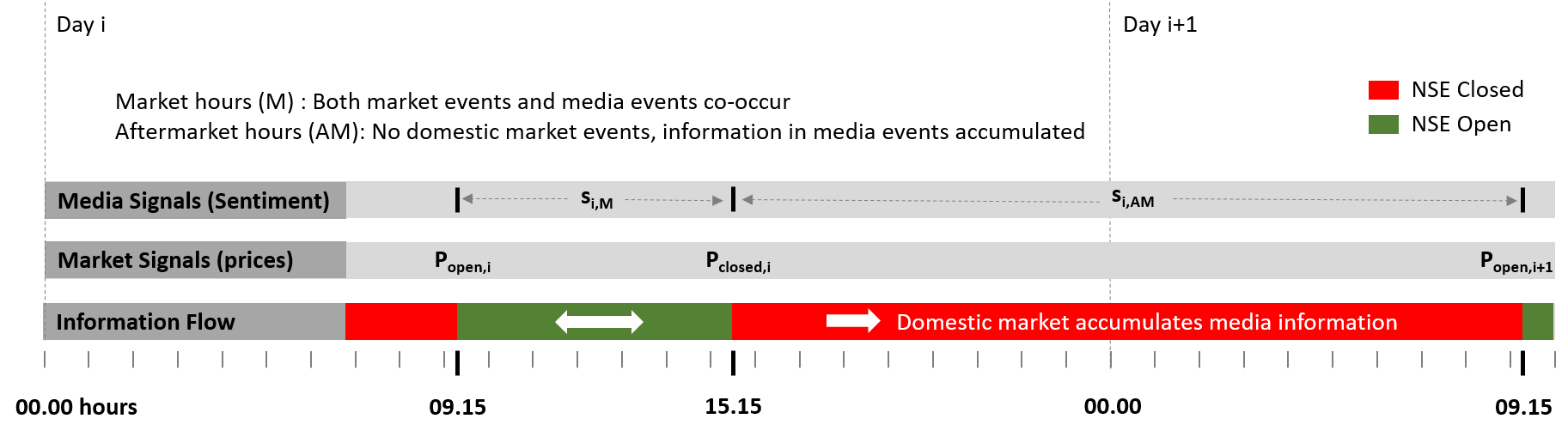}
    \caption{The interactions between market and media over the duration of a open market day indicate that the day can be divided into two durations - market hours and after-market hours. In the market hours, there is two-way exchange of information across the market and the news media. However, in the after-market hours, there are no domestic market events and therefore there is only way of information flow which is the media capturing information from other markets and their events. The information captured by the news media in the after-market hours is likely to reflect in the opening prices of the assets the next open market day.}
    \label{fig:mkt_experiment}
\end{figure*}

The purpose of the experiment is to identify whether the information derived from news flow has predictive power in terms of the ability to affect price changes/updates. As shown in Figure~\ref{fig:mkt_experiment}, the National Stock Exchange (NSE) is open for the duration 09:15 hrs to 15:30 hrs, during which there is a bi-directional information flow between the domestic market and domestic media\footnote{\url{https://www.nseindia.com/resources/exchange-communication-holidays}}. However, during the after-market hours, since the market is closed, the information flow is uni-directional from the market events happening on other exchanges to the domestic media, and there are no domestic market events. The novel information accumulated during the after-market period, is reflected in the corresponding price update on the opening hours of the next market day. We therefore construct the experiment to measure the predictive power of the news flow during the after-market period.  To capture the nature of the relationship over long periods, we considered a 6 year duration from 2012 to 2017, which corresponds to 3 peaks and 4 troughs in the 30-day moving average of the Nifty 50 index. Overall, 175,575 news headlines were utilized for the experiment, which led to 212,828 instances of news text after the entity recognition step. Therefore, we utilized over 210,000 entity-sentiment predictions for the experiment.

\subsection{Variables}

\begin{enumerate}
    \item {Sentiment scores:
    \\We utilize the best performing RoBERTa-based learning scheme discussed in Section-\ref{sec:results}, to get the time-stamped sentiments for the entities present in the news headlines. We adjust for the changes in the composition of the NSE 500 index, for all years from 2012 - 2017. The final sentiment scores are calculated for two durations
    \begin{itemize}
        \item $s_{iM, NSE500}$ - market hours (09:30 AM, ${i}^{th}$ day to 03:30 PM, ${i}^{th}$ day)
        \item $s_{iAM, NSE500}$ - after-market hours (03:30 PM, ${i}^{th}$ day to 09:30 AM, ${i+1}^{th}$ day)
    \end{itemize}}
    \item {Daily log price after-market returns:
    \\We calculate the daily after-market log price returns for the NSE 500 index, using the opening and closing prices as,
    
    \[ d_{i, NSE500} = \log{ \frac{P_{open,i+1,NSE500}}{P_{close,i,NSE500}}} \]
    \[ d_{i, NSE500} = \log{P_{open,i+1,NSE500}} - \log{P_{close,i,NSE500}} \]}
\end{enumerate}

\subsection{Hypothesis}
We hypothesize that the sentiment information derived from the news flow in the after-market hours has a causal effect on the daily after-market price returns. We use a linear regression model to test our hypothesis,
\[ d_{i, NSE500} = \alpha + \beta * s_{iAM, NSE500} + \epsilon \]
with $d_{i, NSE500}$, and $s_{iAM, NSE500}$ as defined above.
%where,
%\\ $d_{i, NSE500}$ - after-market log price returns on $i^{th}$ day,
%\\ $S_{iAM, NSE500}$ - sentiment score of after-market hours on ${i-1}^{th}$ day %\\
The null and alternate hypotheses are as follows,\\
\textbf{$H_{0}$}: On a certain day, the sentiment score of the after market hours during the previous day has no effect on the log price after-market returns with respect to the previous day ($\beta$ = 0) \\
\textbf{$H_{A}$}: On a certain day, the sentiment score of the after-market hours during the previous day has a significant effect on the log price after-market returns with respect to the previous day ($\beta \neq 0$)

\subsection{Results and Discussion}
The results of the regressions are presented in Table \ref{tab:sentiment-analysis}. For the primary sentiment measure, we observe that there exists a statistically significant relation between the log price after-market returns and the sentiment for the after-market duration across the years 2014-17 at a 10\% significance level. For the alternate sentiment measure, we observe that there exists a statistically significant relation between the log price after-market returns and the sentiment for the after-market duration across all the years at a 10\% significance level. On an average we also observe that the alternate sentiment measure is able to capture the variations in NSE 500 index prices more effectively than the primary sentiment measure. In Appendix~\ref{sec:appendix-economicvalue}, we also present an analysis with percentage price returns as the dependent variable. 

Overall, we notice that the information derived from unstructured news flow has predictive value over longer time horizons and also that the predictive value holds across multiple years. While there seems to be economic value in the news information, trading based on such information may not be economically viable if we consider the transaction costs.

% \usepackage{graphicx}
\begin{table*}[htb]
\caption{Results depicting the economic value of sentiments across the years 2012-2017. The primary and alternate sentiment measures show significant effects across the years 2014-2017. The alternate sentiment measure takes into account the neutral sentiments and captures more variation than the primary sentiment measure. In doing so, the alternate sentiment measure exhibits significant effects in the years 2012 and 2013.}
\label{tab:sentiment-analysis}
\resizebox{\textwidth}{!}{%
\begin{tabular}{llllll}
 &  & \multicolumn{2}{c}{Primary sentiment measure ($s_{AM}^{1}$)} & \multicolumn{2}{c}{Alternate sentiment measure ($s_{AM}^{2}$)} \\ \hline
Year & \#observations & beta & r-squared & beta & r-squared \\ \hline
\multicolumn{1}{l|}{2012} & \multicolumn{1}{l|}{261} & 0.00065 & \multicolumn{1}{l|}{0.0073} & 0.00136* & 0.0108 \\
\multicolumn{1}{l|}{2013} & \multicolumn{1}{l|}{260} & 0.00104 & \multicolumn{1}{l|}{0.01} & 0.00242** & 0.0159 \\
\multicolumn{1}{l|}{2014} & \multicolumn{1}{l|}{260} & 0.00242*** & \multicolumn{1}{l|}{0.1093} & 0.00504*** & 0.1304 \\
\multicolumn{1}{l|}{2015} & \multicolumn{1}{l|}{249} & 0.00105** & \multicolumn{1}{l|}{0.0166} & 0.00192** & 0.0168 \\
\multicolumn{1}{l|}{2016} & \multicolumn{1}{l|}{260} & 0.00181** & \multicolumn{1}{l|}{0.0175} & 0.00385** & 0.0206 \\
\multicolumn{1}{l|}{2017} & \multicolumn{1}{l|}{260} & 0.00076* & \multicolumn{1}{l|}{0.0145} & 0.00142* & 0.0139 \\ \hline
\multicolumn{6}{l}{* - p-value \textless 0.1; ** - p-values \textless 0.05; *** - p-value \textless 0.01} \\ \hline
\end{tabular}%
}
\end{table*}

In Appendix~\ref{sec:appendix-var}, we also present a vector autoregression (VAR) based analysis of the daily sentiments and NSE 500 index log price returns across the duration 2012-17 to understand the lags and leads in the time series correlations.

\section{Conclusion and Future Work}
\label{sec:future-work}
Financial sentiment analysis has a rich history of research with specialized dictionaries and datasets. However, most of the studies till date addressed the problem of extracting a single sentiment from the news headlines. One of the key impediments to multiple sentiment extraction is the lack of availability of datasets. To our best knowledge, there are no datasets that provide annotations for multiple entities and sentiments. We therefore make publicly available SEntFiN 1.0 containing 10,753 news headlines with 14,404 entity-sentiment annotations. Using SEntFiN, we address the task of extracting entity-relevant sentiments in a general setting where multiple entities are present in the news headline.

Through our work we provide an comparison of lexicon-based and pre-trained approaches for sentence representations. We definitely conclude that deep bidirectional pre-trained language models such as BERT fine-tuned to SEntFiN outperforms all other learning schemes by a significant margin. We also note that lexicon-based approaches utilizing domain specific task related knowledge are at and above par with generic deep-learning based approaches. Future work can explore the experimentation with varied feature and model architecture choices to better understand the contribution of features and models. We encourage researchers to experiment with other model architectures that are gaining prominence such as GRUs, and other pre-trained language models such as GPT3. Future work can also explore the usage of lexicon-based sentence representations with deep learning architectures to evaluate explainable AI. In our work, we made a conscious design choice to not utilize methods such as dependency parsing to limit our model complexity and focus on the final goal of evaluating the economic value of sentiments. However, the use of graph-based dependency parsing holds promise for extracting contextually relevant phrases and we consider it to be a potential future research direction.

We utilize the model to extract sentiments from news headlines for over 900 Indian companies, and evaluate the economic value of the sentiments at an aggregate market level. Our findings show that the sentiments accumulated in the after-market period have an effect on the market opening prices. We motivate researchers to consider our results and evaluate sentiment-driven trading strategies, and the implications of such a strategy. While we have considered only companies and sectors as entities, we urge future researchers to extend our work and datasets to entities such as commodities like gold and silver \cite{sinha2020impact}, currencies, futures, and other financial entities. 

Our work is grounded in sentiment analysis of news headlines and as such, our methods are focused on the analysis of short text. In order to build comprehensive financial sentiment analysis systems, we urge researchers to consider the task of financial sentiment analysis on large text such as news articles and press releases. The task will entail the use of methods such as coreference resolution, which is a difficult task given that the current publicly available NER systems are not accurate in the financial context. A dictionary-based approach that we have utilized in the study served the study purpose of evaluating economic value of sentiments. However, such an approach is not scalable towards a generalized financial sentiment analysis system, with the ever changing nature of financial markets and growing number of financial entities across geographies. We suggest that future researchers consider this research gap and develop financial domain specific language processing systems and libraries. Such financial NER systems can encourage research in other areas of financial text analysis such as question answering, summarization, robo-advisory. Even with such text processing methods in place, it is likely that the annotated corpora for sentiment analyses on large texts might be required due to the structural complexity and changing themes across the text. Topic models combined with entity-aware sentiment analysis approaches are likely to be effective for large financial text.

\section*{Acknowledgement}
Ankur Sinha would like to acknowledge India Gold Policy Centre (IGPC) for supporting this study under grant number 1815012. He would also like to thank Brij Disa Centre for Data Science and Artificial Intelligence at Indian Institute of Management Ahmedabad for supporting this study.

\bibliographystyle{apacite}
\bibliography{main}

\section{Appendix}

\subsection{Entity-Sentiment Extraction Results}
\label{sec:mean-sd-ml-experiments}

In Table~\ref{tab:average-ml-results} and Table~\ref{tab:sd-mlresults}, we present the average and standard deviation of the performance metrics Accuracy and F1-Score for all the learning schemes across the 31 runs. 

% \usepackage{graphicx}
\begin{table*}[h]
\caption{Average of the performance metrics across the 31 runs.}
\label{tab:average-ml-results}
\resizebox{\textwidth}{!}{%
\begin{tabular}{lllllll}
\hline
 & \multicolumn{2}{c}{positive} & \multicolumn{2}{c}{negative} & \multicolumn{2}{c}{neutral} \\ \hline
Learning Scheme & accuracy & f1-score & accuracy & f1-score & accuracy & f1-score \\ \hline
\multicolumn{1}{l|}{LPS+SVM} & 81.92\% & 75.09\% & 85.95\% & 79.00\% & 77.93\% & 71.73\% \\
\multicolumn{1}{l|}{LPS+GBM} & 82.61\% & 75.99\% & 86.47\% & 80.29\%  & 78.16\% & 71.34\%\\
\multicolumn{1}{l|}{UBT+SVM} & 84.16\% & 79.16\% & 87.70\% & 81.18\% & 80.13\% & 73.25\%\\
\multicolumn{1}{l|}{UBT+GBM} & 84.54\% & 79.87\% & 87.57\% & 81.64\% & 80.93\% & 73.18\%\\
\multicolumn{1}{l|}{UBT+MLP} & 82.15\% & 76.78\% & 87.20\% & 75.25\% & 79.20\% & 73.10\% \\
\multicolumn{1}{l|}{GloVe+RNN} & 83.56\% & 76.23\% & 86.46\% & 74.21\% & 79.01\% & 71.75\% \\
\multicolumn{1}{l|}{GloVe+BiRNN} & 83.03\% & 75.44\% & 85.78\% & 72.76\% & 78.29\% & 71.20\% \\
\multicolumn{1}{l|}{GloVe+LSTM} & 84.73\% & 78.04\% & 87.08\% & 76.10\% & 80.51\% & 74.05\% \\
\multicolumn{1}{l|}{GloVe+BLSTM} & 84.73\% & 78.27\% & 86.75\% & 75.59\% & 80.79\% & 74.26\% \\
\multicolumn{1}{l|}{finBERT} & 90.93\% & 87.02\% & 92.63\% & 85.88\% & 88.87\% & 85.72\%\\ \hline
\end{tabular}%
}
\end{table*}

% \usepackage{graphicx}
\begin{table*}[!htbp]
\caption{Standard deviation of the performance metrics for 31 runs}
\label{tab:sd-mlresults}
\resizebox{\textwidth}{!}{%
\begin{tabular}{lllllll}
\hline
 & \multicolumn{2}{c}{positive} & \multicolumn{2}{c}{negative} & \multicolumn{2}{c}{neutral} \\ \hline
Learning Scheme & accuracy & f1-score & accuracy & f1-score & accuracy & f1-score \\ \hline
\multicolumn{1}{l|}{LPS+SVM} & 0.80\% & 1.30\% & 0.98\% & 1.46\% & 0.80\% & 0.96\%\\
\multicolumn{1}{l|}{LPS+GBM} & 0.75\% & 1.15\% & 0.83\% & 1.17\% & 0.81\% & 1.04\%\\
\multicolumn{1}{l|}{UBT+SVM} & 0.97\% & 1.42\% & 0.86\% & 1.37\% & 0.96\% & 1.07\%\\
\multicolumn{1}{l|}{UBT+GBM} & 0.91\% & 1.23\% & 0.84\% & 1.22\% & 1.11\% & 1.42\%\\
\multicolumn{1}{l|}{UBT+MLP} & 0.75\% & 1.21\% & 0.90\% & 1.24\% & 1.05\% & 1.35\% \\
\multicolumn{1}{l|}{GloVe+RNN} & 0.64\% & 1.12\% & 0.54\% & 0.92\% & 0.63\% & 1.54\% \\
\multicolumn{1}{l|}{GloVe+BiRNN} & 0.32\% & 1.00\% & 0.50\% & 0.82\% & 0.42\% & 1.02\% \\
\multicolumn{1}{l|}{GloVe+LSTM} & 0.36\% & 0.75\% & 0.38\% & 1.62\% & 0.44\% & 0.83\% \\
\multicolumn{1}{l|}{GloVe+BLSTM} & 0.50\% & 0.85\% & 0.38\% & 0.80\% & 0.40\% & 0.83\% \\
\multicolumn{1}{l|}{finBERT} & 0.20\% & 0.28\% & 0.15\% & 0.30\% & 0.20\% & 0.27\%\\ \hline
\end{tabular}%
}
\end{table*}

\subsection{Economic value of sentiments - Percentage price returns}
\label{sec:appendix-economicvalue}

Alternatively, we calculate the daily after-market percentage price returns for the NSE 500 index, using the opening and closing prices as,
    
    \[ d_{i, NSE500} = \frac{P_{open,i+1,NSE500}-P_{close,i,NSE500}}{P_{close,i,NSE500}} \]

The results are presented in Table~\ref{tab:sentiment-analysis-2}. On an average, the alternate sentiment measure which accounts for neutral sentiments expresses higher explanatory power as measured by multiple $R^{2}$, and a stronger significant relationship with percentage price returns across all the years considered.

% \usepackage{graphicx}
\begin{table*}[!htbp]
\caption{Results depicting the economic value of sentiments across the years 2012-2017. The primary and alternate sentiment measures show significant effects across the years 2014-2017. The alternate sentiment measure takes into account the neutral sentiments and captures more variation than the primary sentiment measure. In doing so, the alternate sentiment measure exhibits significant effects in the years 2012 and 2013 as well.}
\label{tab:sentiment-analysis-2}
\resizebox{\textwidth}{!}{%
\begin{tabular}{llllll}
 &  & \multicolumn{2}{c}{Primary sentiment measure ($s_{AM}^{1}$)} & \multicolumn{2}{c}{Alternative sentiment measure ($s_{AM}^{2}$)} \\ \hline
Year & \#observations & beta & r-squared & beta & r-squared \\ \hline
\multicolumn{1}{l|}{2012} & \multicolumn{1}{l|}{261} & 0.00150 & \multicolumn{1}{l|}{0.0073} & 0.00314* & 0.0108 \\
\multicolumn{1}{l|}{2013} & \multicolumn{1}{l|}{260} & 0.00240 & \multicolumn{1}{l|}{0.01} & 0.00557** & 0.0158 \\
\multicolumn{1}{l|}{2014} & \multicolumn{1}{l|}{260} & 0.00557*** & \multicolumn{1}{l|}{0.1088} & 0.00116*** & 0.1298 \\
\multicolumn{1}{l|}{2015} & \multicolumn{1}{l|}{249} & 0.00240** & \multicolumn{1}{l|}{0.0166} & 0.00441** & 0.0168 \\
\multicolumn{1}{l|}{2016} & \multicolumn{1}{l|}{260} & 0.00414** & \multicolumn{1}{l|}{0.0177} & 0.00882** & 0.0209 \\
\multicolumn{1}{l|}{2017} & \multicolumn{1}{l|}{260} & 0.00176* & \multicolumn{1}{l|}{0.0145} & 0.00326* & 0.0139 \\ \hline
\multicolumn{6}{l}{* - p-value \textless 0.1; ** - p-values \textless 0.05; *** - p-value \textless 0.01} \\ \hline
\end{tabular}%
}
\end{table*}

\subsection{Vector Autoregression analysis of sentiments and market index timeseries}
\label{sec:appendix-var}

Vector autoregression (VAR) models are stochastic process models utilized to evaluate the relationships between multi-variate time series. VAR models have risen to prominence in economics and natural sciences \cite{qin2011rise} leading to data-driven theories. In the current context, the dependent variable time-series is the daily NSE500 index percentage price returns ($d_{NSE500}$), and the independent time-series is the daily sentiment measure ($s$). The VAR($p_{1}$, $p_{2}$) model can be elaborated as:

    \[ d_{NSE500}(t) = \alpha_{1} * d_{NSE500}(t-1) + \alpha_{2} * d_{NSE500}(t-2) + ... + \alpha_{p_{1}} * d_{NSE500}(t-p_{1}) \]
    \[                 + \beta_{1} * s(t-1) + \beta_{2} * s(t-2) + ... + \beta_{p_{2}} * s(t-p{2}) + \varepsilon \]

In our analysis, we test for $p_{1}$ = 0, 1, 2, 3 and $p_{2}$ = 0, 1, 2, 3, for the years 2012 to 2017. For sentiment, we utilize the alternate sentiment measure which in our experiments expressed higher explanatory power compared to the primary sentiment measure. In the Table`\ref{tab:varsummary} below we present the models which expressed statistical significance for at least one independent variable. The VAR models were implemented in R using the \textit{dynlm} package and the standard errors were calculated using the \textit{coeftest} package. The results show the $\alpha$, $\beta$ corresponding to the lagged time-series with values rounded to 4 decimal digits. For all the models results, kindly contact the corresponding author.

We observe that the lags in sentiments and index returns differ in their effects across the years. In 2012 and 2014, sentiments lagged by two days showed positive and negative statistically significant relationships with the daily percentage price returns respectively. In 2013, sentiments lagged by one day showed negative and statistically significant relationship with the daily percentage price returns. In 2015, lowest BIC value model indicates that the daily percentage price returns is affected by the sentiments of 1 day and 2 day lags, however, with differing relationships positive and negative respectively. For the years 2016 and 2017, we found that no models exhibited statistical significance. 

\begin{table*}[h]
\centering
\caption{VAR model summary across the years 2012 to 2017}
\label{tab:varsummary}
\begin{tabular}{lllllllll}
\hline
\textbf{year} & \textbf{\parbox{1cm}{VAR}} & \textbf{$d_{t-1}$} & \textbf{$d_{t-2}$} & \textbf{$d_{t-3}$} & \textbf{$s_{t-1}$} & \textbf{$s_{t-2}$} & \textbf{$s_{t-3}$} & \textbf{BIC} \\ \hline
\multicolumn{1}{l|}{2012} & \multicolumn{1}{l|}{(0,2)} &  &  &  & 0.0014 & 0.0018+ &  & -2596.30 \\
\multicolumn{1}{l|}{2012} & \multicolumn{1}{l|}{(0,3)} &  &  &  & 0.0015 & 0.0019+ & -0.0016 & -2582.01 \\
\multicolumn{1}{l|}{2012} & \multicolumn{1}{l|}{(1,2)} & 0.0777 &  &  & 0.0012 & 0.0017+ &  & -2592.31 \\
\multicolumn{1}{l|}{2012} & \multicolumn{1}{l|}{(3,1)} & 0.0837 & 0.1244+ & -0.0788 & 0.0013 &  &  & -2578.15 \\
\multicolumn{1}{l|}{2012} & \multicolumn{1}{l|}{(1,3)} & 0.0877 &  &  & 0.0013 & 0.0017+ & -0.0017 & -2578.43 \\
\multicolumn{1}{l|}{2012} & \multicolumn{1}{l|}{(2,3)} & 0.078 & 0.1147 &  & 0.0014 & 0.0015 & -0.0019 & -2576.32 \\
\multicolumn{1}{l|}{2013} & \multicolumn{1}{l|}{(3,0)} & -0.0886 & -0.0514 & -0.1293+ &  &  &  & -1242.86 \\
\multicolumn{1}{l|}{2013} & \multicolumn{1}{l|}{(3,1)} & -0.0892 & -0.052 & -0.1275+ & -0.0001+ &  &  & -1238.07 \\
\multicolumn{1}{l|}{2013} & \multicolumn{1}{l|}{(3,2)} & -0.0904 & -0.0531 & -0.129+ & -0.0001+ & -0.0001 &  & -1233.33 \\
\multicolumn{1}{l|}{2013} & \multicolumn{1}{l|}{(3,3)} & -0.0895 & -0.0518 & -0.1278+ & -0.0001+ & -0.0001 & 0.0001 & -1228.70 \\
\multicolumn{1}{l|}{2014} & \multicolumn{1}{l|}{(3,0)} & 0.2252** & -0.0242 & 0.1558+ &  &  &  & -2601.36 \\
\multicolumn{1}{l|}{2014} & \multicolumn{1}{l|}{(2,0)} & 0.2287** & 0.0085 &  &  &  &  & -2609.64 \\
\multicolumn{1}{l|}{2014} & \multicolumn{1}{l|}{(1,0)} & 0.23** &  &  &  &  &  & -2626.18 \\
\multicolumn{1}{l|}{2014} & \multicolumn{1}{l|}{(1,1)} & 0.27*** &  &  & -0.0016+ &  &  & -2624.99 \\
\multicolumn{1}{l|}{2014} & \multicolumn{1}{l|}{(2,2)} & 0.269*** & 0.0189 &  & -0.0016* & 0.0002 &  & -2603.08 \\
\multicolumn{1}{l|}{2014} & \multicolumn{1}{l|}{(1,2)} & \multicolumn{2}{l}{0.2734***} &  & -0.0016* & 0.0003 &  & -2608.55 \\
\multicolumn{1}{l|}{2014} & \multicolumn{1}{l|}{(3,1)} & 0.2621*** & -0.0087 & 0.1589+ & -0.0016* &  &  & -2600.38 \\
\multicolumn{1}{l|}{2014} & \multicolumn{1}{l|}{(3,2)} & 0.2621*** & -0.0086 & 0.1589+ & -0.0016* & 0 &  & -2594.83 \\
\multicolumn{1}{l|}{2014} & \multicolumn{1}{l|}{(3,3)} & 0.2623*** & -0.0168 & 0.1727* & -0.0016+ & 0.0001 & -0.0005 & -2589.76 \\
\multicolumn{1}{l|}{2014} & \multicolumn{1}{l|}{(1,3)} & 0.271*** &  &  & -0.0016+ & 0.0003 & 0.0001 & -2593.63 \\
\multicolumn{1}{l|}{2014} & \multicolumn{1}{l|}{(2,3)} & 0.266*** & 0.0218 &  & -0.0016+ & 0.0002 & 0.0001 & -2588.19 \\
\multicolumn{1}{l|}{2015} & \multicolumn{1}{l|}{(3,0)} & 0.0455 & 0.0235 & -0.1368 &  &  &  & -2423.70 \\
\multicolumn{1}{l|}{2015} & \multicolumn{1}{l|}{(2,2)} & 0.0275 & 0.0438 &  & 0.0019+ & -0.0021** &  & -2432.36 \\
\multicolumn{1}{l|}{2015} & \multicolumn{1}{l|}{(1,2)} & 0.0281 &  &  & 0.0019* & -0.002* &  & -2437.43 \\
\multicolumn{1}{l|}{2015} & \multicolumn{1}{l|}{(3,1)} & 0.018 & 0.0167 & -0.146+ & 0.0017+ &  &  & -2421.83 \\
\multicolumn{1}{l|}{2015} & \multicolumn{1}{l|}{(3,2)} & 0.027 & 0.0479 & -0.1396+ & 0.0021* & -0.002* &  & -2421.53 \\
\multicolumn{1}{l|}{2015} & \multicolumn{1}{l|}{(3,3)} & 0.0173 & 0.0522 & -0.1252 & 0.0021* & -0.0018* & -0.0009 & -2417.08 \\
\multicolumn{1}{l|}{2015} & \multicolumn{1}{l|}{(1,3)} & 0.0148 &  &  & 0.002* & -0.0017* & -0.0013+ & -2423.52 \\
\multicolumn{1}{l|}{2015} & \multicolumn{1}{l|}{(2,3)} & 0.0136 & 0.0507 &  & 0.002* & -0.0018* & -0.0013+ & -2418.62 \\
\multicolumn{1}{l|}{2015} & \multicolumn{1}{l|}{(0,2)} &  &  &  & 0.002* & -0.002* &  & -2442.79 \\
\multicolumn{1}{l|}{2015} & \multicolumn{1}{l|}{(0,3)} &  &  &  & 0.002* & -0.0017* & -0.0013+ & -2429.02 \\ \hline
\multicolumn{9}{l}{+ : p-value \textless 0.1; * : p-value \textless 0.05; ** : p-value \textless 0.01; *** p-value \textless 0.001} \\ \hline
\end{tabular}
\end{table*}

\end{document}